%% file: main.tex
\documentclass[10pt,twocolumn,letterpaper]{article}

\usepackage[pagenumbers]{cvpr} 
\usepackage[ruled,linesnumbered]{algorithm2e}
\usepackage{multirow}
\usepackage{bbding}
\input{preamble}

\definecolor{cvprblue}{rgb}{0.21,0.49,0.74}
\usepackage[pagebackref,breaklinks,colorlinks,citecolor=cvprblue]{hyperref}
\usepackage{colortbl}
\usepackage{booktabs}

\def\paperID{17278} 
\def\confName{CVPR}
\def\confYear{2024}

\title{Mixed-Precision Quantization for Federated Learning on Resource-Constrained Heterogeneous Devices}

\author{Huancheng Chen\\
University of Texas at Austin\\
{\tt\small huanchengch@utexas.edu}
\and
Haris Vikalo\\
University of Texas at Austin\\
{\tt\small hvikalo@ece.utexas.edu}
}

\begin{document}
\maketitle
\input{sec/0_abstract}    
\input{sec/1_intro}

\input{sec/2_related_work}

\input{sec/3_methodology}

\input{sec/4_experiments}
\input{sec/5_conclusion}
{
    \small
    \bibliographystyle{ieeenat_fullname}
    \bibliography{main}
}


\end{document}

%% file: preamble.tex
\usepackage[dvipsnames]{xcolor}


%% file: sec/0_abstract.tex
\begin{abstract}
While federated learning (FL) systems often utilize quantization to battle communication and computational bottlenecks, they have heretofore been limited to deploying fixed-precision quantization schemes. Meanwhile, the concept of mixed-precision quantization (MPQ), where different layers of a deep learning model are assigned varying bit-width, remains unexplored in the FL settings. We present a novel FL algorithm, FedMPQ, which introduces mixed-precision quantization to resource-heterogeneous FL systems. Specifically, local models, quantized so as to satisfy bit-width constraint, are trained by optimizing an objective function that includes a regularization term which promotes reduction of precision in some of the layers without significant performance degradation. The server collects local model updates, de-quantizes them into full-precision models, and then aggregates them into a global model. To initialize the next round of local training, the server relies on the information learned in the previous training round to customize bit-width assignments of the models delivered to different clients. In extensive benchmarking experiments on several model architectures and different datasets in both iid and non-iid settings, FedMPQ outperformed the baseline FL schemes that utilize fixed-precision quantization while incurring only a minor computational overhead on the participating devices.
\end{abstract}

%% file: sec/1_intro.tex
\section{Introduction}
\label{intro}
\emph{Federated Learning} (FL) \citep{fedavg,fedprox,fednova,scaffold} paradigm enables collaboration among numerous distributed devices (clients) while protecting their privacy by avoiding collection of data stored at those devices. The vanilla FL algorithm, FedAvg \citep{fedavg}, deploys a client-server framework in which the server periodically collects locally trained/updated models from the clients and aggregates them into the global model. A number of of follow-up studies \citep{convergence1,convergence2,convergence3,convergence4} explored convergence guarantees for FedAvg with convex and non-convex loss functions in the settings where clients' data is independent and identically distributed (i.i.d.). However, in real-world scenarios, the participating clients' data is likely to be non-i.i.d., which has detrimental effects on the convergence properties of FedAvg \citep{convergence_noniid}. This motivated a considerable amount of research aiming to address the challenges of statistical heterogeneity in federated learning \citep{fedprox, scaffold, fednova, moon,feddyn, fedmd, pFL, pFLsurvey, dfkd, fedhkd,feddpms,hicsfl}. On a related note, clients in large-scale FL systems are likely to have varying communication and computational resources at their disposal, requiring development of more sophisticated, constraints-aware schemes \citep{Advances}. Recently, there has been a growing interest in compressing models so that they can be deployed on devices having limited resources \citep{compression}. A particularly effective model compression technique 
is that of model quantization \citep{uveqfed,fedpaq,constraint, adaquant, bitwidth, quped}.

The use of model quantization in FL has primarily been motivated by the high communications costs that arise when the server collects updates from a potentially very large number of clients. Reducing the precision of local updates by using lower bit-width to represent model parameters allows the clients with limited bandwidth to efficiently communicate with the server \citep{fedpaq, uveqfed, constraint,sattler2019robust, jhunjhunwala2021adaptive, haddadpour2021federated,adaquant}. However, existing quantization-based FL algorithms expect each client to learn a full-precision local model regardless of the potential constraints on the client's computational resources. In real-world settings, clients such as smart phones or wearable devices may not have sufficient amount of memory to allow full-precision model training. Nevertheless, there are only few prior studies considering quantization-aware training in FL \citep{bitwidth,quped,abdelmoniem2021towards,fedqnn,elkordy2022heterosag}, all of them implementing \emph{fixed-precision quantization} (FPQ) of local models. The recent advancements in \emph{mixed-precision quantization} (MPQ) \citep{wu2018mixed,haq, hawq,bsq, fracbits,uhlich2019mixed,abdelmoniem2021towards} remain unexplored in the federated learning settings. Adapting the existing MPQ schemes to federated learning is far from straightforward. The existing MPQ schemes typically train a full-precision model to enable computing the quantization errors which are then used to optimize layer-wise bit-width allocation. Both search-based \citep{haq,wu2018mixed} and optimization-based \citep{fracbits,ddq, uhlich2019mixed} MPQ methods consume significantly more computation than the FL training itself. 

The aim of this paper is to introduce mixed-precision quantization to federated learning, developing an efficient framework that addresses the challenge of resource heterogeneity in FL. In particular, we study federated learning system where the clients deploy local models at average bit-width that may vary from one device to another. In such scenarios, clients with low average bit-width budgets cannot run computationally intensive MPQ methods. We propose FedMPQ, a novel \textbf{Fed}erated learning algorithm with \textbf{M}ixed-\textbf{P}recision \textbf{Q}uantization, which enables training of quantized local models within the allocated bit-width budget. FedMPQ first initializes local models as fixed-precision quantized networks that satisfy clients' average bit-width budget, and then converts these quantized networks into a representation that allows bit-level sparsity-promoting training. In particular, while learning local models whose parameters admit binary representation, the clients deploy a group Lasso regularization term which imposes a trade-off between the task loss and bit-sparsity. The precision of layers that end up having parameters which exhibit higher degree of sparsity is reduced to allow increasing precision of other layers. During the aggregation step, FedMPQ employs the \emph{pruning-growing} strategy where the server aggregates clients' models (locally trained at potentially different bit-widths), resulting in the global model. Before transmitting the global model to a client, the bit-width of the model is adjusted to match the client's bit-width budget. To evaluate the effectiveness of FedMPQ, we conducted experiments  on CIFAR10, CIFAR100 and Tiny-Imagenet \citep{tiny}. FedMPQ outperforms the baseline in non-i.i.d. data settings and achieves performance similar to the FPQ8 baseline even though a subset of the clients train their models at a very low precision. The contributions of the paper are summarized as follows:
\begin{itemize}
    \item  We propose a method for mixed-precision quantization in FL which does not require training full-precision models on devices.
    \item We introduce a \emph{pruning-growing} strategy for allocating layer-wise bit-width without a need for computationally expensive procedures that may violate resource constraints.
    \item We conduct extensive experiments in non-i.i.d. FL settings where the clients have heterogeneous computational resources, demonstrating the performance of the proposed FedMPQ. 
\end{itemize}

%% file: sec/2_related_work.tex
\section{Related Work}
\label{related_work}
\subsection{Quantization for Federated Learning}  

Much of the existing research on federated learning under resource constraints focuses on quantizing local updates for the purpose of reducing communication bandwidth \citep{bitwidth,adaquant}. The milestone work, FedPAQ \citep{fedpaq}, presents a federated learning framework where each client communicates quantized local updates to the server, and provides analytical convergence guarantees for both strongly-convex and non-convex settings. FedCOMGATE \citep{fedcomgate} extends the ideas of FedPAQ to introduce a local gradient tracking scheme mitigating detrimental effect introduced when learning from non-i.i.d. data. UVeQFed \citep{uveqfed} takes a step further, utilizing vector quantization to move from lossy to lossless compression. The follow-up studies \citep{constraint,jhunjhunwala2021adaptive,mao2022communication,adaquant} propose adaptive quantization for local updates in communication-constrained settings while still requiring clients to locally train full-precision models. 

AQFL \citep{abdelmoniem2021towards} took a step towards mitigating the detrimental effects of computational heterogeneity by training quantized models with bit-widths proportional to the clients' computational resources. The follow-up works \citep{elkordy2022heterosag,chen2021dynamic, bitwidth} developed a series of methods improving the server's aggregation of local updates quantized at varying levels. However, these methods are limited to fixed-precision quantization, assigning the same bit-width to the entire model. This motivates us to explore mixed-precision quantization in FL schemes as an alternative approach to learning in computationally heterogeneous scenarios.
 
\subsection{Mixed-Precision Quantization}    

Aiming to enable stronger expressiveness of learned models, mixed-precision quantization (MPQ) assigns different bit-widths to different layers/modules of the models. The MPQ strategies, which generally attempt to assign bit-widths in proportion to the importance of different layers, can be organized in three categories: (1) search-based, (2) optimization-based, and (3) metric-based. 

Search-based methods such as HAQ \cite{haq} and AutoQ \citep{autoq} utilizes reinforcement learning (RL) \citep{sutton2018reinforcement} to pursue optimal bit-widths where the model performance is set as the reward. DNAS \citep{wu2018mixed} and SPOS\citep{guo2020single} apply neural architecture search (NAS) to explore the quantization space which is growing exponentially with the number of layers. Since both RL-based and NAS-based methods require an enormous amount of computation, it is unrealistic to implement them in FL settings.

Optimization-based methods approach MPQ from an optimization perspective by formulating the bit-width allocation problem using differentiable variables and applying the straight-through \citep{ste, uhlich2019mixed,fracbits, ddq} or gumbel-softmax \citep{gumbel,EdMIPS,sdq,hmq} estimator. However, this leads to mixed-integer programming problems which are NP-hard, rendering the use of optimization-based methods in FL scenarios practically infeasible.

In contrast to the search-based and optimization-based methods, metric-based methods leverage a variety of metrics to evaluate the importance of layers and subsequently decide on the bit-width allocation. Such metrics include the eigenvalues of the Hessian matrix \citep{hawq,hawqv2,hawqv3,chen2021towards}, orthogonality conditions \citep{ompq}, entropy measures \cite{sun2022entropy}, synaptic flow \citep{synflow} and learnable layer-wise importance \citep{importance}. The computation of the aforementioned metrics is relatively expensive as it requires the information from full-precision models. Recently, BSQ \citep{bsq} presented a bit-pruning MPQ strategy that achieves high compression rate while preserving model performance; however, BSQ simulates binary representation with floating-point values, making it impractical for FL settings.

%% file: sec/3_methodology.tex
\section{Methodology}
\label{method}
\subsection{Federated Learning with Quantization}
In a cross-device scenario with $N$ clients, where client $n$ owns a private dataset $\mathcal{D}_{n}$, the standard federated learning (FL) considers training a single global model $\mathbf{W}$ by minimizing the loss (empirical risk)
\begin{equation}
    \min_{\mathbf{W}}\mathcal{L}(\mathbf{W}) = \sum_{n=1}^{N}p_{n}\mathcal{L}_{n}(\mathbf{W}),
\end{equation}
where $\mathcal{L}_{n}(\cdot)$ is the local loss function on $\mathcal{D}_{n}$ and $p_{n} \in [0, 1]$ denotes the weight assigned to client $n$. At round $t$ of FedAvg \citep{fedavg}, a widely used FL algorithm, each participating client locally trains a model $\mathbf{W}_{n}^{t}$ on local data and communicates it to the server; the server aggregates the collected local models to form the global model $\mathbf{W}^{t} = \sum_{k=1}^{N} p_{n}\mathbf{W}_{n}^{t}$. Since the clients with restricted resources may not implement full-precision model training, one may instead opt for quantization-aware training. In that case, the aggregation at the server can be described as
\begin{equation}
\label{aggregation}
\begin{aligned}
    \mathbf{W}^{t} &= \sum_{k=1}^{N}p_{n}Q_{\mathbf{b}_{n}}(\mathbf{W}_{n}^{t})\\
    &\text{s.t.} \quad \mathbf{b}_{n} \cdot \mathbf{m}/ \Vert\mathbf{m}\Vert_{1} \leq v_{n},  \forall n \in [N],
\end{aligned}
\end{equation}
where $Q_{\mathbf{b}_{n}}$ denotes the mixed-precision quantizer, $\mathbf{b}_{n} \in \mathbb{Z}^{L}$ denotes the bit-width assigned to each layer; $\mathbf{m} = \{M^{(1)}, \dots, M^{(L)}\} \in \mathbb{Z}^{L}$ is the number of parameters in each layer of model, $L$ is the number of layers, and $v_{n}$ is the budget of the average bit-width for client $n$. Since in resource-heterogeneous FL $v_{n}$ varies across clients, naive aggregation according to Eq.~\ref{aggregation} may discard beneficial knowledge of high-precision models. 

\subsection{Binary Representation of Model Parameters}
\label{bit}

In the conventional deep neural networks (DNNs), the full-precision model parameters are typically stored in $32$-bit floating-point form. Compared to the floating-point format, integer-arithmetic operations such as \emph{mult}, \emph{add} and \emph{shift}, on model parameters in fixed-point representation are more efficient and hardware-friendly. Following studies \citep{qat,bitwidth,bsq, sekikawa2022bit}, we assume clients perform training using low-precision fixed-point quantization. A $B$-bit matrix $\mathbf{W}^{(l)} \in \mathbb{R}^{C\times K}$ of the parameters in the $l$-th layer of the model can be represented in binary format with a ternary matrix $\mathbf{B}^{(l)} \in \{0,1\}^{  B \times  C\times K  }$, a layer-wise scaling factor $s^{(l)}$ with floating-point value, and a layer-wise zero-point $z^{(l)} \in \mathbb{Z}^{+}$ as
\begin{equation}
\label{forward}
    \mathbf{W}_{j,k}^{(l)} = \frac{s^{(l)}}{2^{B} - 1} \left(\sum_{i = 1}^{B}2^{i - 1}  \mathbf{B}_{i,j,k}^{(l)}  -  z^{(l)}\right),
\end{equation}
where $z^{(l)}$ is typically set to $2^{B-1}$ (signed integer); the scaling factor $s^{(l)}$ is updated at each training round according to the maximum absolute value of the parameters at the $l$-th layer. Therefore, the product between activation $\mathbf{A} \in \mathbb{R}^{K \times U}$ and $ \mathbf{W}^{(l)}$ can be simplified by using \emph{shift} and \emph{add} according to
\begin{equation}
\begin{aligned}
    \mathbf{A}_{\cdot, u}^{\top}\cdot \mathbf{W}_{j, \cdot}^{(l)}   &= \frac{s^{(l)}}{2^{B} - 1}  \sum_{k=1}^{K}\mathbf{A}_{k,u}\left(\sum_{i=1}^{B}2^{i-1}  \mathbf{B}_{i,j,k}^{(l)} - z^{(l)}\right)\\
    &= \frac{s^{(l)}}{2^{B} - 1}   \left(\sum_{i=1}^{B}2^{i-1} \mathbf{A}_{\cdot, u}^{\top}  \cdot \mathbf{B}_{i, j, \cdot}^{(l)} -  z^{(l)}\sum_{k=1}^{K} \mathbf{A}_{k,u}\right).
\end{aligned}
\end{equation}

Note that $\mathbf{B}^{(l)}$ consists of discrete values $0$ and $1$, which cannot be searched for via gradient descent. To optimize over these binary parameters, we adopt the straight-through estimator (STE) \citep{ste} as in \citep{bsq}. STE enables a quantized network to forward pass intermediate signals using model parameters represented in fixed-point format (as shown in Eq.~\ref{forward}) while computing the gradients with continuous floating-point parameters as
\begin{equation}
\label{ste_pass}
\begin{aligned}
    &\textbf{Forward: } \mathbf{W}_{j,k}^{(l)} = \frac{s^{(l)}}{2^{B} - 1} \left(\sum_{i = 1}^{B}2^{i - 1}  \mathbf{B}_{i,j,k}^{(l)}  -  z^{(l)}\right)\\
    &\textbf{Backward: } \frac{\partial	\mathcal{L}}{\partial \mathbf{B}_{i,j,k}^{(l)}} = \frac{s^{(l)} 2^{i-1}}{2^{B} - 1} \frac{\partial	\mathcal{L}}{\partial \mathbf{W}_{j,k}^{(l)}}
\end{aligned}
\end{equation}
where \textbf{Backward} pass follows from the chain rule. BSQ \citep{bsq} relaxes the binary constraint and allows using floating-point values to update $\mathbf{B}_{i,j,k}^{(l)}$, which means that BSQ trains the networks with simulated quantization \citep{integer}. Different from BSQ, we adapt the WAGE \citep{integer} strategy and update binary parameters using integer operations via the \emph{power-of-two} function $S(\cdot)$ defined as
\begin{equation}
    S(x) = 2^{\lceil \log x \rfloor},
\end{equation}
where $S(x)$ returns the nearest power-of-two of $x$. Then we compute an update of $\mathbf{W}_{j,k}^{(l)}$ via gradient descent with step size $\eta$ according to
\begin{equation}
\begin{aligned}
 \Delta \mathbf{W}_{j,k}^{(l)} &= - \nu \cdot \frac{s^{(l)}}{2^{B} - 1} \sum_{i = 1}^{B}2^{i - 1}   S\left(\eta \left | \frac{\partial	\mathcal{L}}{\partial \mathbf{B}_{i,j,k}^{(l)}} \right |\right)\\
 &= - \nu \cdot \frac{s^{(l)}}{2^{B} - 1} \sum_{i = 1}^{B}2^{q_{i} - 1},
\end{aligned}
\end{equation}
where $\nu \in \{-1,1\}$ denotes the sign of $\frac{\partial	\mathcal{L}}{\partial \mathbf{W}_{j,k}^{(l)}}$, and $q_{i}$ is the power of $2^{i} \cdot S\left(\eta \left | \frac{\partial	\mathcal{L}}{\partial \mathbf{B}_{i,j,k}^{(l)}} \right |\right)$. According to Eq.~\ref{ste_pass}, $q_{i}$ is strictly ascending, i.e., $q_{i} < q_{j}$ for $i < j$. Let $\mathcal{Q} = \{q_1, \dots, q_{B}\}$ denote the set of $q_{i}$, 
and let $\max(\mathcal{Q})$ be the maximum element in $\mathcal{Q}$. On one hand, if $\max(\mathcal{Q}) > B$ the absolute value of $\Delta \mathbf{W}_{j,k}^{(l)}$ exceeds the scale $s^{(l)}$ and thus $\Delta \mathbf{W}_{j,k}^{(l)}$ needs to be clipped as
\begin{equation}
\label{clip1}
    C(\Delta \mathbf{W}_{j,k}^{(l)}) = - \nu \cdot \frac{s^{(l)}}{2^{B} - 1} \sum_{i = 1}^{B}2^{i - 1}.
\end{equation}
Note that $\Delta \mathbf{W}_{j,k}^{(l)}$ cannot be represented by a fixed-point integer value if $q_{i} \leq 0 \in \mathcal{Q}$. Adapting the strategy of updating parameters with small-magnitude gradients in WAGE \citep{integer}, $\Delta \mathbf{W}_{j,k}^{(l)}$ is converted to
\begin{equation}
\label{clip2}
\begin{aligned}
    C(\Delta \mathbf{W}_{j,k}^{(l)}) =& - \nu \cdot \frac{s^{(l)}}{2^{B} - 1}\sum_{i = 1}^{B}2^{i - 1}\cdot \mathcal{I}\{i \in \mathcal{Q}\} \\
    & - \nu \cdot \frac{s^{(l)}}{2^{B} - 1}\text{Bernoulli}(\sum_{q_{i} \leq 0}2^{q_{i} - 1}),
\end{aligned}
\end{equation}
where $\mathcal{I}(\cdot)$ is an indicator, and $\text{Bernoulli}(\cdot)$ randomly samples decimal parts to either $0$ or $1$. When the magnitude of the gradient is small, the integer part of $\Delta \mathbf{W}_{j,k}^{(l)}$ is always $0$, which impedes the update of the parameters. Due to the second term on the right hand side in Eq.~\ref{clip2}, $\mathbf{W}_{j,k}^{(l)}$ is updated with the minimum step size even if the gradient is very small. After converting $\Delta \mathbf{W}_{j,k}^{(l)}$ to the fixed-point format, one can update the parameters according to
\begin{equation}
\label{update}
\mathbf{W}_{j,k}^{(l)} \xleftarrow{} \text{Clipping } (\mathbf{W}_{j,k}^{(l)} + C(\Delta \mathbf{W}_{j,k}^{(l)}), \text{min}^{(l)}, \text{max}^{(l)}),
\end{equation}
where $\text{min}^{(l)} =  -  \frac{s^{(l)}}{2^{B} - 1} z^{(l)}$ denotes the minimum value of the parameters, and $\text{max}^{(l)} =  \frac{s^{(l)}}{2^{B} - 1} \left(2^{B} - 1  -  z^{(l)}\right)$ is the maximum value of the parameters in the $l$-th layer. Since the updated $\mathbf{W}_{j,k}^{(l)}$ is in fixed-point format, updating binary representation $\mathbf{B}_{i,j,k}^{(l)}$ is straightforward.

\subsection{Sparsity-Promoting Training}
\label{sparsity}
\begin{figure}[t]
\begin{center}
\includegraphics[width= 1 \linewidth]{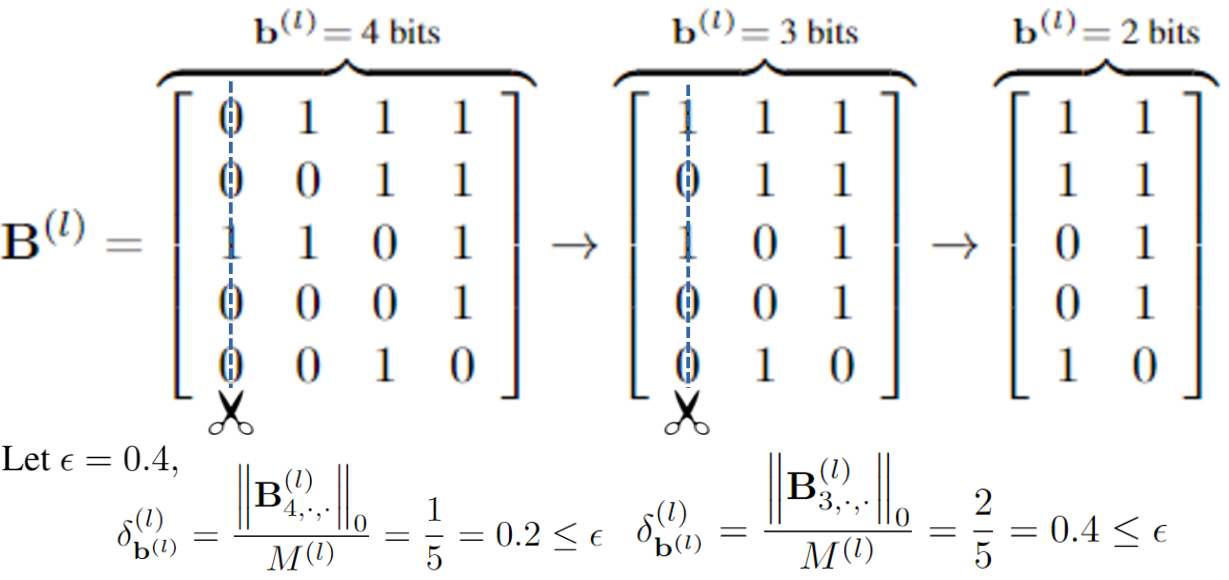}
\end{center}
\vspace{-0.15 in}
   \caption{An example of pruning the MSBs in the model parameters $\mathbf{B}^{(l)}$. Since the $4$-th and $3$-rd bits in $\mathbf{B}^{(l)}$ have a fraction of $1$'s below the threshold $\epsilon$, the bit-width $\mathbf{b}^{(l)}$ of the $l$-th layer is reduced from $4$ bits to $2$ bits. For clarity, $\epsilon$ is set to $0.4$ in the example but the value of $\epsilon$ is smaller in the experiments.} 
\label{pruning}
\end{figure}

Sparsity-promoting regularizers including L1 (Lasso) and L2 (ridge) have been widely used to induce sparsity e.g. during feature selection. Following BSQ \citep{bsq}, we use group Lasso regularization \citep{lasso} to promote obtaining highly sparse parameters and ensure stable convergence. The group Lasso regularizer for the parameters in the $l$-th layer is defined as
\begin{equation}
    R_{\text{GL}}(\mathbf{B}^{(l)}) = \sum_{i = 1}^{\mathbf{b}^{(l)}}\left\Vert \mathbf{B}^{(l)}_{i,\cdot,\cdot}\right\Vert_{2},
\end{equation}
where $\mathbf{b}^{(l)}$ denotes the bit-width of the $l$-th layer, and $\mathbf{B}^{(l)}_{i,\cdot,\cdot} \in \{0,1\}^{C\times K}$ is the $i$-th binary representation position of the parameters in the $l$-th layer. Since the number of parameters typically differ from one layer to another, we vary the weights assigned to the group Lasso regularizers according to the memory constraints. Specifically, the objective function used in local training is formulated as
\begin{equation}
\label{local_traing}
\mathcal{L}_{\text{local}} = \mathcal{L}_{\text{task}}(\mathbf{B}^{(1:L)}) + \lambda \sum_{l = 1}^{L}\frac{M^{(l)}}{M} R_{\text{GL}}(\mathbf{B}^{(l)}),
\end{equation}
where $M^{(l)}$ denotes the number of parameters in the $l$-th layer, $M = \sum_{l=1}^{L}M^{(l)}$, and $\lambda$ is a non-negative pre-determined hyper-parameter. Note that $\mathcal{L}_{\text{task}}(\mathbf{B}^{(1:L)})$ can be computed through the STE forward pass while the gradients can be computed through the STE backward pass as indicated in Eq.~\ref{ste_pass}. Let $\delta^{(l)}_{i}$ denote the sparsity of the $i$-th binary representation position in the $l$-th layer,
\begin{equation}
    \delta^{(l)}_{i} = \left\Vert \mathbf{B}^{(l)}_{i,\cdot,\cdot}\right\Vert_{0}/M^{(l)}.
\end{equation}
Essentially, $\delta^{(l)}_{i} $ is the proportion of the parameters having $1$ at the $i$-th binary representation position. The smaller $\delta^{(l)}_{i}$ is, the higher the sparsity at the $i$-th position of the parameters' binary representation. We set up a threshold $\epsilon$ and prune the most significant bits (MSBs) if $\delta^{(l)}_{\mathbf{b}^{(l)}} \leq \epsilon$ (see Figure~\ref{pruning}).

\subsection{The End-to-End Training Procedure}
The local sparsity-promoting training and MSBs pruning results in reduced bit-widths of the model parameters in certain layers, leading to a higher compression rate while maintaining the model's performance. However, our aim is to leverage mixed-precision quantization to improve the performance of the global model given all of the available resources, rather than pursue high compression rate. To this end, we propose a \emph{pruning-growing} procedure at the server that restores the average bit-width allocated to each local model. In particular, the server aggregates the collected local models characterized by different bit-width allocations into the global model $\mathbf{W}^{t+1}$ with the global bit-width allocation $\mathbf{b}^{t+1}$. For a given client, $\mathbf{b}^{t+1}$ is adjusted to satisfy the client's bit-width constraint via $2$ operations: (1) \emph{pruning}: reducing the bit-width of a subset of layers for the clients with low bit-width budget until the constraint in Eq.~\ref{aggregation} is satisfied; (2) \emph{growing}: increasing the bit-width of a subset of layers for the clients with high bit-width budget until violating the constraint in Eq.~\ref{aggregation}. Alg.~\ref{alg2} specifies the greedy policy used to select the layers for pruning or growing. Next, we provide an outline of the end-to-end training procedure, as formalized in Alg.\ref{alg1}.

\subsubsection{Initialization and Local Training}
\label{initialization}
At the start of training, the local bit-width assignment of client $n$ is initialized as $\hat{\mathbf{b}}_{n}^{0} = \{v_{n}, \dots ,v_{n}\}$, while the global model $\mathbf{W}^{0}$ (full precision) is randomly initialized at the server. Then, $\mathbf{W}^{0}$ is converted to a $v_{n}$-bit fixed-point representation $\mathbf{G}_{n}^{0}$, with scale $\mathbf{s}_{n}^{0} \in \mathbb{R}^{L}$ set to the maximum absolute value of the model parameters in each layer, i.e.,
\begin{equation}
    s_{n}^{(l),0} = \max_{j,k} \left |\mathbf{W}^{(l),0}_{j,k}\right|.
\end{equation}
The customized global model $\mathbf{G}_{n}^{0}$ is then sent to client $n$ to be used for initialization of local training. With such an initialization, the local model $\mathbf{B}_{n}^{0} \xleftarrow{} \mathbf{G}_{n}^{0}$ satisfies the constraint of average bit-width $v_{n}$.

\input{sec/algorithm1}

At global round $t$, client $n$ receives the global model $\mathbf{G}_{n}^{t}$ and uses it to initialize $\mathbf{B}_{n}^{t}$. The client $n$ updates $\mathbf{B}_{n}^{t}$, $\mathbf{b}_{n}^{t}$ and potentially $\mathbf{s}_{n}^{t}$, and possibly prunes MSBs of the model parameters across different layers as discussed in Sections~\ref{bit} and \ref{sparsity}. After updating, the client $n$ sends $\mathbf{B}_{n}^{t+1}$ and $\mathbf{b}_{n}^{t+1}$ to the server for aggregation.

\subsubsection{Models Aggregation}
The server collects local models $\mathbf{B}_{n}^{t+1}$ and the corresponding local bit-width assignments $\mathbf{b}_{n}^{t+1}$ from the participating clients, $n \in [N]$, and then converts the collected fixed-point models to the floating-point models according to Eq.~\ref{forward}. The global model and the average bit-width are computed according to  Eq.~\ref{aggregation},
\begin{equation}
\label{aggreation_server}
\mathbf{W}^{t+1} = \sum_{n=1}^{N} p_{n}\mathbf{W}_{n}^{t+1}, \quad \mathbf{b}^{t+1} = \sum_{n=1}^{N} p_{n}\mathbf{b}_{n}^{t+1},
\end{equation}
where $p_{n} = v_{n} \cdot \left| \mathcal{D}_{n} \right|/P$ is proportional to the number of samples in local dataset $\mathcal{D}_{n}$ and client $n$'s budget $v_{n}$, and $P = \sum_{i = 1}^{N}v_{i} \cdot \left| \mathcal{D}_{i} \right|$. Note that larger weights are assigned to the clients having higher bit-width budgets and training on larger local datasets.

\input{sec/algorithm2}

The local bit-width assignment $\mathbf{b}_{n}^{t+1}$ is learned during the local sparsity-promoting training, where the bit-widths for less sensitive layers are reduced while the bit-widths for more sensitive layers are preserved. Note that the sensitivity of layers can vary across clients with non-i.i.d data, as it is affected by the data distribution. $\mathbf{b}^{t+1}$ aggregates local bit-width assignments and is reflective of the importance of different layers. 

\subsubsection{Post-Aggregation Adjustment}
Due to the constraints on the local bit-width, the aggregated full-precision global model $W^{t+1}$ cannot be directly broadcasted to the clients. Similar to the initialization described in Section~\ref{initialization}, the server needs to customize different fixed-point global models $\mathbf{G}_{n}^{t+1}$ with bit-width assignments $\hat{\mathbf{b}}_{n}^{t+1}$ based on $\mathbf{b}^{t+1}$ and budget $v_{n}$. There are three options for updating $\mathbf{b}_{n}^{t+1}$: 
\begin{equation}
    \hat{\mathbf{b}}^{t+1} = \left\{
    \begin{aligned}
        &\mathbf{b}^{t+1}, \text{ if } v= v_{n}, \\
        &\text{pruning}(\mathbf{b}^{t+1}), \text{ if } v > v_{n}, \\
        &\text{growing}(\mathbf{b}^{t+1}), \text{ if } v < v_{n}, \\
    \end{aligned}
    \right.
\end{equation}
where $v = \mathbf{b}^{t+1} \cdot \mathbf{m}/ \left\Vert\mathbf{m}\right\Vert_{1}$. The first assignment is handled in a straightforward manner. For the later two we apply a \emph{greedy} policy for executing pruning or growing. Specifically, when $v > v_{n}$, we attempt to reduce the bit-width of the layer having the most parameters, attempting to satisfy the budget constraint $v_{n}$ while maintaining the precision of other layers. When $v < v_{n}$, we first increase the bit-width of the layer with the fewest parameters. If two layers have the same number of parameters, the pruning preference is given to the layer whose bit-width in the last round of local sparsity-promoting training had been reduced more, as that suggests this layer is less important. It is beneficial to record the bit-width change during local training by defining $\Delta \mathbf{b}_{n}^{t} = \hat{\mathbf{b}}_{n}^{t} - \mathbf{b}_{n}^{t+1}$. Further details of pruning and growing are specified in Alg.~\ref{alg2}.

%% file: sec/algorithm1.tex
\begin{algorithm}
\small
 \SetAlgoLined
  \KwIn{global round $T$, local epochs $\tau$, budget $v_{n}$, local data $\mathcal{D}_{n}$, number of parameters $\mathbf{m}$, $\lambda$ and $\epsilon$.  }
  \KwOut{the global model $\mathbf{W}^{T}$}
  initialize $\mathbf{W}^{0}$, $\mathbf{G}_{n}^{0}$, $\hat{\mathbf{b}}_{n}^{0} \xleftarrow{} \{v_{n}, \dots, v_{n}\}, \forall n \in [N]$ \;
  \For{t = 0, \dots, T}{
    \tcc{\textcolor{blue}{local training in the clients}}
    \For{n = 1, \dots, N}{
        $\mathbf{B}_{n}^{t} \xleftarrow{} \mathbf{G}_{n}^{t}$, $\mathbf{b}_{n}^{t} \xleftarrow{} \hat{\mathbf{b}}_{n}^{t}$\;
        $\mathbf{B}_{n}^{t+1}, \mathbf{b}_{n}^{t+1} \xleftarrow{} \text{LocalUpdate}(\mathcal{D}_{n}, \tau, \lambda, \epsilon)$\;
        send $\mathbf{B}_{n}^{t+1}, \mathbf{b}_{n}^{t+1}$ to the server\;
     }
     \tcc{\textcolor{blue}{aggregation in the server}}
     \For{n = 1, \dots, N}{
     $\mathbf{W}_{n}^{t+1} \xleftarrow{} \text{ConvertToFP}(\mathbf{B}_{n}^{t+1}, \mathbf{b}_{n}^{t+1})$\;
     }
     $\mathbf{W}^{t+1} \xleftarrow{} \sum_{n}^{N}p_{n}\mathbf{W}_{n}^{t+1}$, $\mathbf{b}^{t+1} \xleftarrow{} \sum_{n}^{N}p_{n}\mathbf{b}_{n}^{t+1}$\;
     \tcc{\textcolor{blue}{post-aggregation adjustment}}
     \For{n = 1, \dots, N}{
     $\Delta \mathbf{b}_{n}^{t} \xleftarrow{} \hat{\mathbf{b}}_{n}^{t} - \mathbf{b}_{n}^{t+1}$\;
     $\hat{\mathbf{b}}_{n}^{t+1} \xleftarrow{} \textbf{Pruning-Growing}(\mathbf{b}^{t+1},\Delta \mathbf{b}_{n}^{t},  \mathbf{m}, v_{n})$\;
     $\mathbf{G}_{n}^{t+1} \xleftarrow{} \text{Binary-Representation}(\mathbf{W}_{n}^{t+1}, \hat{\mathbf{b}}_{n}^{t+1})$\;
     }
    }
    
  \caption{FedMPQ}\label{alg1}
\end{algorithm}

%% file: sec/algorithm2.tex
\begin{algorithm}
\small
 \SetAlgoLined
  \KwIn{$\mathbf{b}^{t+1},\Delta \mathbf{b}_{n}^{t},  \mathbf{m}, v_{n}$}
  \KwOut{bit-width assignment for client $n$, $\hat{\mathbf{b}}_{n}^{t+1}$}
  initial: $\hat{\mathbf{b}}_{n}^{t+1} \xleftarrow{} \mathbf{b}^{t+1}$, $ v \xleftarrow{} \hat{\mathbf{b}}_{n}^{t+1} \cdot \mathbf{m}/ \left\Vert\mathbf{m}\right\Vert_{1}$, $\mathbf{d} \xleftarrow{} \text{argDescending}(\mathbf{m}\odot(\Delta \mathbf{b}_{n}^{t} + \mathbf{1}))$, $cur \xleftarrow{} 0$\;
  \tcc{\textcolor{blue}{Pruning}}
  \While{$v > v_{n}$ \textbf{and}  $cur < |\mathbf{m}|$}{
    $l \xleftarrow{} \mathbf{d}[cur]$ \;
    \eIf{$\hat{\mathbf{b}}_{n}^{t+1}[l] > 1$}{
      $\hat{\mathbf{b}}_{n}^{t+1}[l] \xleftarrow{} \hat{\mathbf{b}}_{n}^{t+1}[l] - 1$, $v \xleftarrow{} v - \mathbf{m}[l]/\left\Vert\mathbf{m}\right\Vert_{1}$\;
      }{
      $cur \xleftarrow{} cur + 1$\;
      }
    }
    $cur \xleftarrow{} |\mathbf{m}| - 1$\;
    \tcc{\textcolor{blue}{Growing}}
    \While{$v < v_{n}$ \textbf{and}  $cur > 0$}{
    $l \xleftarrow{} \mathbf{d}[cur]$ \;
    \eIf{$\hat{\mathbf{b}}_{n}^{t+1}[l] < 8$}{
      $\hat{\mathbf{b}}_{n}^{t+1}[l] \xleftarrow{} \hat{\mathbf{b}}_{n}^{t+1}[l] + 1$, $v \xleftarrow{} v + \mathbf{m}[l]/\left\Vert\mathbf{m}\right\Vert_{1}$\;
      }{
      $cur \xleftarrow{} cur - 1$\;
      }
    }
  \caption{Pruning-Growing}\label{alg2}
\end{algorithm}

%% file: sec/4_experiments.tex
\section{Experiments}
\label{experiments}
\subsection{Setup}
\label{setup}
We evaluate the performance of the proposed FedMPQ in various settings on three datasets: CIFAR10, CIFAR100 and Tiny-ImageNet \citep{le2015tiny}. We perform the experiments using ResNet20 \citep{resnet} model for CIFAR10/100, and ResNet44 \citep{resnet} model for Tiny-ImageNet. We use the mini-batch stochastic gradient descent (SGD) with a learning rate initialized to $0.1$ in all experiments. The SGD momentum is set to $0.9$ and the weight decay is set to $0.0005$. The batch size is set to $64$ and the number of global rounds is set to $50$, with $5$ local epochs within each global round. The value of the bit-pruning threshold $\epsilon$ is $0.03$, while the regularizing hyper-parameter $\lambda$ is equal to $0.01$. The number of clients is $10$; unless stated otherwise, the fraction of participating clients is $0.5$. Following the strategy in \citep{yurochkin2019bayesian}, we use Dirichlet distribution with varying concentration parameter $\alpha$ to generate data partitions at different levels of heterogeneity (smaller $\alpha$ leads to generating less balanced data). 

Since there exist no prior methods for mixed-precision quantization-aware training (QAT) in FL, we primarily focus on comparing FedMPQ with the AQFL \citep{abdelmoniem2021towards}, the first method to deploy fixed-precision quantization-aware training in FL. As FedMPQ only modifies the precision of the model weights, we fix the precision of the activation to $4$ bits in both FedMPQ and AQFL throughout the entire training process. Furthermore, we implement FedAvg with full-precision (FP32) training, $8$-bits fixed-precision quantization-aware training (FPQ8) \citep{qat}, and two communication-efficient FL methods, FedPAQ \citep{fedpaq} and UVeQFed \citep{uveqfed}, which train full-precision local models but then utilize scalar and vector quantization to compress the local updates before communicating them to the server. 

\subsection{Effects of Data Heterogeneity}
\input{sec/table1}
To evaluate our method in the scenarios characterized by varied levels of data heterogeneity, we conduct $3$ sets of experiments where $\alpha$ takes on values from 
$\{0.1,0.5,1\}$; these correspond to severely imbalanced, moderately imbalanced and mildly imbalanced data, respectively. Results of the experiments are reported in Table~\ref{table1}. As can be seen there, performance of the global model of all the considered methods deteriorates as the data heterogeneity increases. FedPAQ and UVeQFed, the two communication-efficient FL approaches, achieve performance comparable to the FP32 baseline when $\alpha = 0.5$ and $1$ but experience performance degradation when $\alpha = 0.1$. A significant performance decline is experienced in the experiments with FPQ8 due to the low capacity of the low-precision models. However, the performance gap between FP32 and FPQ8 narrows as the level of data heterogeneity increases. For instance, when $\alpha = 1$, the test accuracy of FP32 is $11.3 \%$ higher than that of FPQ8; when $\alpha = 0.1$, the test accuracy difference is $5.8 \%$ (experiments on CIFAR10). When the data is extremely imbalanced, FedAvg suffers from the so-called ``client-drift" problem \citep{scaffold} caused by overfitting on the local data. As a result, the model capacity advantage of FP32 (due to having higher precision) might not make as much of a positive impact on the accuracy, explaining the narrowing gap between FP32 and FPQ8.
\input{sec/table2}

Since local models with ultra-low precision are aggregated into the global model, performance of AQFL shows further deterioration. The proposed method, FedMPQ outperforms AQFL (implementing fixed-precision quantization) in all scenarios even though training models under the same resource constraints. As shown in Table.\ref{table1}, FedMPQ outperforms AFQL at most $9.1\%$, $8.2 \%$ and $2.9\%$ test accuracy on CIFAR10, CIFAR100 and Tiny-ImageNet respectively. On CIFAR10/100 datasets, FedMPQ nearly preserves performance of FPQ8 baseline in the settings $\alpha = 0.5$ and $1$ by efficiently allocating precision to different layers, even though training the global model on resource-constrained heterogeneous devices.

\subsection{Scalability}
To evaluate the effect of the system size on the performance of FedMPQ, we conducted $3$ sets of experiments on CIFAR10/100 data in the FL system with $10, 20$ and $40$ clients. To simulate a resource-constrained heterogeneous system, we allocate to $20\%, 20\%,20\%$ and $30\%$ clients the bit-width budget of $2, 4, 6$ and $8$ bits, respectively. The concentration parameter $\alpha$ is set to $0.5$ and kept constant throughout the experiments. 

Not surprisingly, as the results in Table~\ref{table2} show, the performance of all schemes  deteriorates as the FL training involves an increasingly larger number of clients with ultra-low budget. FedMPQ consistently outperforms AQFL, with up to $9.1\%$ and $8.2\%$ higher accuracy on CIFAR10 and CIFAR100, respectively. Note that FedMPQ manages to approach the performance of FPQ8 in the settings involving $10$ clients, though the gap (as expected) widens as the number of clients participating in the system grows.

\subsection{The Number of Local Epochs}

\begin{figure}[t] 
    \centering
	  \subfloat[CIFAR10]{
       \includegraphics[width=0.49\linewidth]{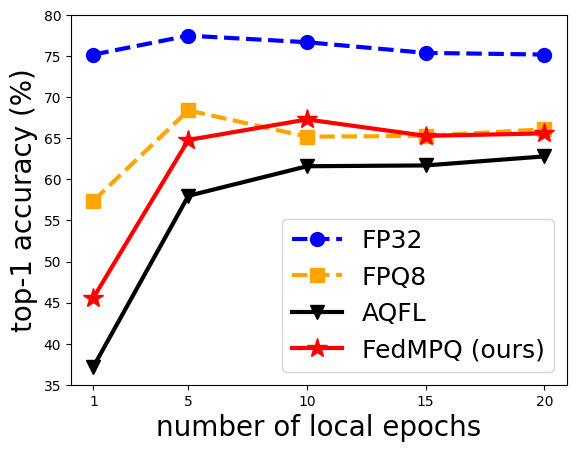}}
	  \subfloat[CIFAR100]{
        \includegraphics[width=0.49\linewidth]{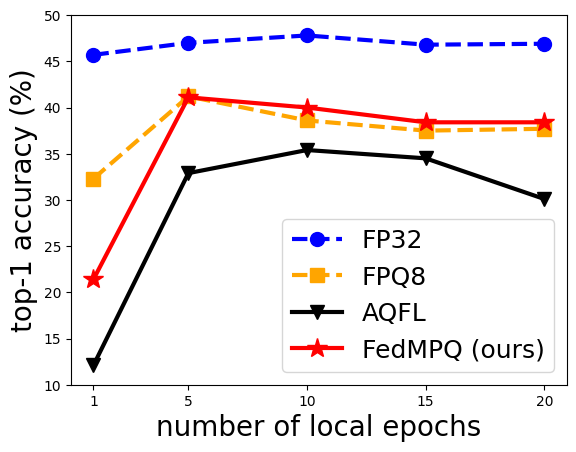}}
	  
	\caption{Top-1 accuracy vs. the number of local epochs. In all experiments, the number of clients is $N=10$, the concentration parameter $\alpha=0.5$, and the number of global rounds is set to $50$. All results are the final test accuracy after $50$ global rounds.}
\label{local_epoch} 
\end{figure}

We study the impact of the number of local epochs on the system's performance by conducting $5$ sets of experiments where the number of local epochs is varied over $\{1,5,10,15,20\}$. The results are shown in Figure~\ref{local_epoch}. As can be seen there, when the number of local epochs is set to $1$, the three quantization-aware training methods show significant performance degradation while the impact on FP32 baseline is only minor. This is likely due to underfitting of the quantized models. For larger numbers of local epochs, FedMPQ provides a global model whose performance is comparable to FPQ8, demonstrating remarkable improvement over AQFL.

\subsection{Fine-Tuning Hyper-Parameters}
The experiments discussed in the previous section are conducted with hyperparameters $\epsilon = 0.03$ and $\lambda = 0.01$. Note that $\epsilon$ and $\lambda$ jointly affect the training: $\lambda$ controls the weight of the group Lasso regularization used in local training while $\epsilon$ controls the threshold for pruning the MSBs after local training. To explore the space of hyper-parameters, we consider a number of configurations with varied values of $\epsilon \in \{0.01,0.02,0.03,0.04,0.05\}$ and $\lambda \in \{0.1,0.01,0.001\}$. As the results shown in Table~\ref{table3} indicate, when the threshold $\epsilon$ is too large, the accuracy considerably deteriorates since a larger fraction of model parameters gets compressed. Selecting large regularization weights, e.g. $\lambda = 0.1$, leads to the performance drop since FedMPQ focuses on pursuing higher bit-level sparsity. Our experiments suggest that to achieve satisfactory performance, one should select $\epsilon \leq 0.03$ and $\lambda \leq 0.01$.

\input{sec/table3}
\subsection{An Ablation Study}
In this section, we empirically analyze the effect of each procedure in FedMPQ by a comparison to the AQFL baseline. The three procedures that distinguish FedMPQ from AQFL include: (1) the group Lasso regularization in the objective function as described in Section~\ref{sparsity}; (2) bit-level pruning in the most significant bits (MSBs); and (3) Alg.~\ref{alg2} which restores the precision of local models. We refer to different combinations of these procedures, shown in Table~\ref{table4}, as settings (1)-(5).

According to the results for setting (1) in Table~\ref{table4}, the group Lasso regularization achieves high performance while promoting bit-level sparsity, with small $1.2\%$ and $0.3\%$ drops in accuracy on CIFAR10 and CIFAR100, respectively. As shown in the results for setting (2), MSBs pruning without sparsity-promoting training causes more severe performance degradation, with  $2.9\%$ and $4.2\%$ accuracy drop on these two datasets. Finally, by combining the group Lasso regularization with MSBs pruning (setting (3)) enables the global model to achieve performance close to the baseline, even though the precision is reduced after MSBs pruning. Interestingly, the group Lasso regularization forces the clients to learn local models with highly sparse binary weight representations, implying no degradation due to bit-pruning. 

Algorithm~\ref{alg2} enables clients to recover layer-wise precision budget allocated to their local models, resulting in significant performance improvements. For instance, test accuracy in setting (4) is $10.3\%$ and $11.4\%$ higher than in setting (2); setting (5) achieves $10.9 \%$ and $9.3\%$ higher accuracy than setting (3) on CIFAR10 and CIFAR100, respectively. 

This ablation study provides an insight in how FedMPQ operates: the group Lasso regularization forces clients to learn a local model with bit-level sparsity while preserving performance; MSBs pruning allows reduction of the precision of local models without major performance degradation; Algorithm~\ref{alg2}, implemented at the server, conducts \emph{pruning-growing} to restore the precision of local models so they fully exploit allocated bit-width budgets while seeking effective bit-width allocations to layers. The results in Table~\ref{table4} suggest that Algorithm~\ref{alg2} plays a major role in helping FedMPQ achieve high accuracy.
\input{sec/table4}

%% file: sec/table1.tex
\begin{table*}[t]
\caption{Test accuracy (\%) of the considered schemes as the concentration parameter $\alpha$ takes values from $\{0.1, 0.5, 1\}$. 
The number of clients in these experiments is $10$, while their average bit-width budgets are $\mathbf{v} = \{2,2,4,4,4,6,6,6,8,8\}$ ($v_{n}$ denotes the budget of client $n$). The numbers in the column ``Update", ``Weight" and ``Activation" indicate the bits used to store the local update values, model weights and the activation signals, respectively. The last column indicates whether the scheme needs to train a full-precision model or not.}
\centering
\small
\begin{tabular}{l|ccccccccc|cccc}
\bottomrule[1pt]
\label{table1}
& \multicolumn{3}{c}{\textbf{CIFAR10}}                         & \multicolumn{3}{c}{\textbf{CIFAR100}}                        & \multicolumn{3}{c|}{\textbf{Tiny-ImageNet}}       
&\multirow{2}{*}{\textbf{Update}}   &\multirow{2}{*}{\textbf{Weight}}  &\multirow{2}{*}{\textbf{Activation}} &\multirow{2}{*}{\textbf{Full-Precision?}} \\
\cline{0-9}
 $\alpha$         & $0.1$ & $0.5$                          & $1$    & $0.1$ & $0.5$  & $1$  & $ 0.1$ & $0.5$  & $1$  &  & & &\\
\hline
\rowcolor[HTML]{EFEFEF} 

FP32            & 60.3                           & 77.5                            & 82.1 & 40.3                    & 47.0                     & 49.6                  & 24.6    & 35.1                     &38.1   & 32                        & 32     & 32  &\CheckmarkBold                   \\
FedPAQ          & 56.3                              & 77.0                              & 81.2                     & 39.7                    & 46.8                     & 48.4                  &22.87     & 34.6                     & 37.5
  &       $v_{n}$                    & 32 & 32 &\CheckmarkBold                       \\
UVeQFed        & 56.8                              & 76.7                              & 81.5                     & 38.6                    & 46.5                     & 48.8                  &21.3     & 34.3& 37.4
  &   $v_{n}$                         & 32          & 32 &\CheckmarkBold               \\
\hline
\rowcolor[HTML]{EFEFEF} 
FPQ8           & 54.5                              & 68.4                              & 70.8 & 35.4                    & 41.3                     & 42.3                  & 23.8    & 33.4                     &35.6   & 8                         & 8     & 8    &\XSolidBrush              \\
AQFL           & 44.3        & 58.0                              & 62.1                     & 23.8                    & 32.9 & 36.1                  &17.1     & 23.5                     &25.3   &                          $v_{n}$ &    $v_{n}$           & 4     &\XSolidBrush        \\
\textbf{FedMPQ} & \textbf{49.1} & \textbf{67.1} & \textbf{69.3}            & \textbf{31.7}           & \textbf{41.1}            & \textbf{43.6}         & \textbf{20.3}     & \textbf{27.0}            & \textbf{28.2}  &                          $v_{n}$ &  $v_{n}$  & 4    &\XSolidBrush \\
\toprule[1pt]
\end{tabular}
\end{table*}

%% file: sec/table2.tex
\begin{table}[]
\caption{Test accuracy (\%) as the number of clients $N$ varies over $10, 20$ and $40$. Here, $20\%$, $30\%$, $30\%$ and $20\%$ clients have average bit-width budget of $2$ bits, $4$ bits, $6$ bits and $8$ bits, respectively. The concentration parameter $\alpha$ is set to $0.5$. }
\centering
\begin{tabular}{l|cccccc}
\bottomrule[1pt]
\label{table2}
 & \multicolumn{3}{c}{\textbf{CIFAR10}}                         & \multicolumn{3}{c}{\textbf{CIFAR100}}  \\
\hline
 $N$      & 10 & 20                         & 40    &  10 & 20  & 40 \\
\hline
\rowcolor[HTML]{EFEFEF} 

FP32            & 77.5                           & 68.1                           & 64.5 &  47.0                   &  40.1                    & 35.0  \\
FedPAQ          & 77.0                              & 61.5                              & 59.9                    &  46.8                   &38.9                 &33.1  \\
UVeQFed        & 76.7                              & 66.4                            & 63.2                    & 46.5              &39.5                     & 34.1 \\
\hline
\rowcolor[HTML]{EFEFEF} 
FPQ8           & 68.4                              & 56.3                              & 48.4 &  41.3                   &36.2                     &31.9  \\
AQFL           & 58.0        & 49.7                              & 37.3                    &  32.9                 &20.4  &13.9   \\
\textbf{FedMPQ} & \textbf{67.1} & \textbf{56.8} & \textbf{45.4}            & \textbf{41.1}          & \textbf{26.1}             & \textbf{19.9}   \\
\toprule[1pt]
\end{tabular}
\end{table}

%% file: sec/table3.tex
\begin{table}[t]
\caption{Test accuracy ($\%$) of FedMPQ running with different combinations of threshold value $\epsilon$ and regularization weights $\lambda$. All experiments are on CIFAR10 with $10$ clients and $\alpha = 0.5$.}
\centering
\begin{tabular}{l|ccccc}
\bottomrule[1pt]
\label{table3}
 & \multicolumn{5}{c}{threshold value $\epsilon$}  \\
\hline
 $\lambda$       & 0.01 & 0.02    & 0.03    &  0.04 & 0.05  \\
\hline

0.1             & 64.1                        & 63.4                          &65.9   & 63.9                  &  64.8                    \\
0.01            &66.4                              &68.0                              &\cellcolor[HTML]{C0C0C0} 67.1                    &64.7                      &64.3                 \\
0.001          & 67.1                            &70.0                           &65.5                   &64.7               & 64.4  \\

\toprule[1pt]
\end{tabular}
\end{table}

%% file: sec/table4.tex
\begin{table}[t]
\caption{Test accuracy ($\%$) of the global model trained using different combinations of the FedMPQ subroutines. ``Lasso" refers to the group Lasso regularization; ``MSBs" denotes bit-level pruning. All experiments involve $10$ clients. $\alpha, \epsilon$ and $\lambda$  are set to $0.5$, $0.03$ and $0.01$, respectively (the same setting as in Section~\ref{setup}). }
\centering
\begin{tabular}{l|cc}
\bottomrule[1pt]
\label{table4}
 & \textbf{CIFAR10} & \textbf{CIFAR100}  \\
\hline
AQFL (baseline)  & 58.0                           &32.9                             \\
(1) Lasso            & 56.8                           &32.6                             \\
(2) MSBs            &55.1                           &28.7                             \\
(3) MSBs + Lasso        & 56.2                               & 31.8                                            \\
(4) MSBs + Alg.\ref{alg2}            &65.4                                &                 40.1                            \\
(5) MSBs + Lasso + Alg.\ref{alg2}  & \textbf{67.1} & \textbf{41.1}  \\
\toprule[1pt]
\end{tabular}
\end{table}

%% file: sec/5_conclusion.tex
\section{Conclusion}
\label{conclusion}
In this paper we presented FedMPQ, a novel framework for heterogeneous, resource-constrained federated learning systems, which aims to judiciously utilize bit-width budget of clients by conducting mixed-precision quantization. A group Lasso regularizer applied in local training promotes sparsity of the binary representation of the model parameters, and then reduces the precision of a subset of the local model's layers. The server deploys a greedy \emph{pruning-growing} procedure that restores precision of the pruned local models to fully exploit the assigned bit-width budget. We conducted extensive experiments on several benchmark datasets for a number of problem configurations that simulate resource-heterogeneity across clients. The experimental results demonstrate that FedMPQ outperforms the fixed-precision quantization baseline, and provides performance remarkably close to the 8-bit quantization baseline.